# Exploring CTC Based End-To-End Techniques for Myanmar Speech Recognition

Khin Me Me Chit[✉] and Laet Laet Lin

University of Information Technology, Yangon, Myanmar
{khinmemechit,laetlaetlin}@uit.edu.mm

**Abstract.** In this work, we explore a Connectionist Temporal Classification (CTC) based end-to-end Automatic Speech Recognition (ASR) model for the Myanmar language. A series of experiments is presented on the topology of the model in which the convolutional layers are added and dropped, different depths of bidirectional long short-term memory (BLSTM) layers are used and different label encoding methods are investigated. The experiments are carried out in low-resource scenarios using our recorded Myanmar speech corpus of nearly 26 h. The best model achieves character error rate (CER) of 4.72% and syllable error rate (SER) of 12.38% on the test set.

## 1 Introduction

ASR plays a vital role in human-computer interaction and information processing. It is the task of converting spoken language into text. Over the past few years, automatic speech recognition approached or exceeded human-level performance in languages like Mandarin and English in which large labeled training datasets are available [1]. However, the majority of languages in the world do not have a sufficient amount of training data and it is still challenging to build systems for those under-resourced languages.

A traditional ASR system is composed of several components such as acoustic models, lexicons, and language models. Each component is trained separately with a different objective. Building and tuning these individual components make developing a new ASR system very hard, especially for a new language. By taking advantage of Deep Neural Network's ability to solve complex problems, end-to-end approaches have gained popularity in the speech recognition community. End-to-end models replaced the sophisticated pipelines with a single neural network architecture.

The most popular approaches to train an end-to-end ASR include Connectionist Temporal Classification [2], attention-based sequence-to-sequence models [3], and Recurrent Neural Network (RNN) Transducers [4].

CTC defines a distribution over all alignments with all output sequences. It uses Markov assumptions to achieve the label sequence probabilities and solves this efficiently by dynamic programming. It has simple training and decoding schemes and showed great results in many tasks [1, 5, 6].

The sequence-to-sequence model contains an encoder and a decoder and it usually uses an attention mechanism to make alignment between input features and output symbols. They showed great results compared to CTC models and in some cases surpassed them [7]. However, their computational complexity is high and they are hard to parallelize.

RNN-Transducer is an extension of CTC. It has an encoder, a prediction network, and a joint network. It uses the outputs of encoder and prediction networks to predict the labels. It is popular due to its capability to do online speech recognition which is the main challenge for attention encoder-decoder models. Due to its high memory requirement in training and the complexity in implementation, there is less research for RNN-Transducer although it has obtained several impressive results [8].

The CTC-based approach is significantly easier to implement, computationally less expensive to train, and produces results that are close to state-of-the-art. Therefore it is a good start to explore the end-to-end ASR models using CTC. In this paper, several experiments are carried out with the CTC based speech models on low-resource scenarios using the Myanmar language dataset ($\sim 26$ h). We compare different label encoding methods: character-level encoding, syllable-level encoding, and sub-word level encoding on our ASR model. We also vary the number of BLSTM layers and explore the effect of using a deep Convolutional Neural Network (CNN) encoder on top of BLSTM layers.

## 2   Related Work

Most of the early works in Myanmar speech recognition are mostly Hidden Markov Model (HMM) based systems. Hidden Markov Model is used together with Gaussian Mixture Model (GMM) and Subspace Gaussian Mixture Model (SGMM) and the performance of the model increases with the increased use of training data [9].

HMM-GMM models are also used in spontaneous ASR systems and tuning acoustic features, number of senones and Gaussian densities tends to affect the performance of the model [10].

Since the Deep Neural Networks (DNNs) have gained success in many areas, they are also used in automatic speech recognition systems usually in the combination with HMMs [11–14]. Research has shown that the DNN, CNN, and Time Delay Network models outperformed the HMM-GMM models [14].

However, many of these works needed several components where we have difficulties to tune each individual component. End-to-end ASR models simplified these components into a single pipeline and achieved state-of-the-art results in many scenarios. To the best of our knowledge, no research has been conducted on end-to-end training in Myanmar ASR systems. In this paper, we introduce an end-to-end Myanmar ASR model with the CTC based approach.

## 3 End-To-End CTC Model

This section introduces the CTC based end-to-end model architecture. We explore the architectures with the initial layers of VGGNet [15] and up to 6 layers of BLSTM. Batch Normalization is applied after every CNN and BLSLM layer to stabilize the learning process of the model. A fully connected layer with the softmax activation function is followed after the BLSTM layers. The CTC loss function is used to train the model. Figure 1 shows the model architecture, which includes a deep CNN encoder and a deep bidirectional LSTM network.

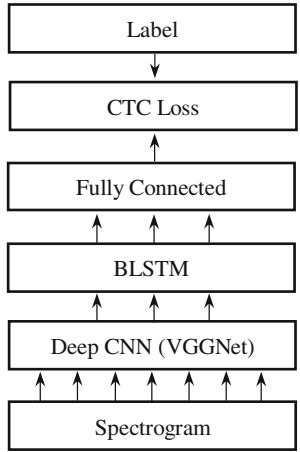

**Fig. 1.** The architecture of CTC based end-to-end ASR model. The different architectures are explored by varying the number of BLSTM layers from 3 to 6 and by removing and adding deep convolutional layers.

### 3.1 Deep Convolutional LSTM Network

Since the input audio features are continuous, the time dimension is usually downsampled to minimize the running time and complexity of the model. In order to perform the time reduction, one way is to skip or concatenate the consecutive time frames. In this work, the time reduction is achieved by passing the audio features through CNN blocks containing two max-pooling layers. It is observed that using convolutional layers speeds up the training time and also helps with the convergence of the model. As an input to the CNN layers, we use two dimensions of spectrogram features and the dimension of length one as the channel. After passing through two max-pooling layers, the input features are downsampled to (1/4 × 1/4) along the time-frequency axes. The architecture of the CNN layers is shown in Fig. 2.

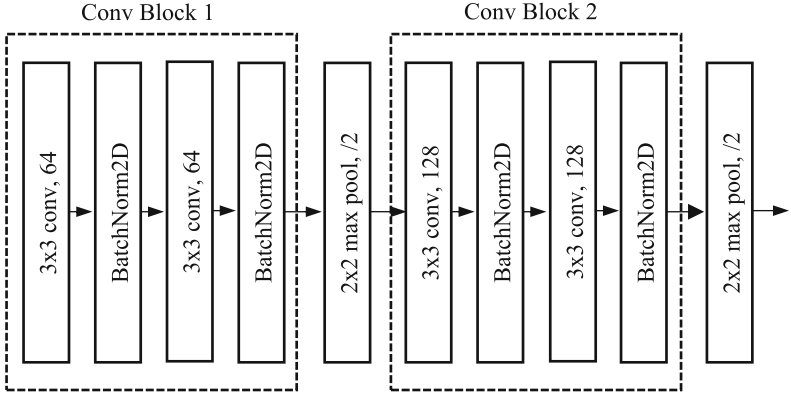

**Fig. 2.** The architecture of convolutional layers (VGGNet). Each convolutional block consists of two convolutional layers with Batch Normalization and a max pooling layer. Each convolutional layer uses the ReLU activation function.

A stack of BLSTM layers is followed after the CNN layers. The models with 3 to 6 layers of BLSTM are experimented using 512 hidden units in each layer and direction. Since the Batch Normalization is shown to improve the generalization error and accelerate training [16], Batch Normalization is added after every BLSTM layer.

### 3.2 Connectionist Temporal Classification (CTC)

In speech recognition, the alignment between input audio and output text tokens is needed to consider. CTC is a way to solve this problem by mapping the input sequence to the output sequence of shorter length. Given the input sequence of audio features $X = [x_1, x_2, \ldots, x_T]$ and the output sequence of labels $Y = [y_1, y_2, \ldots, y_U]$, the model tries to maximize $P(Y|X)$ the probability of all possible alignments that assigns to the correct label sequence. An additional blank symbol is introduced in CTC to handle the repeated output tokens and silence audio. The CTC objective [17] for a $(X, Y)$ pair is described as follows:

$$P(Y|X) = \sum_{A \in A_{X,Y}} \prod_{t=1}^{T} P_t(a_t|X). \quad (1)$$

where $A$ is a set of valid alignments, $t$ denotes a time-step and $a$ indicates a single alignment. In Eqs. 1, we need to solve the expensive computation of the sum of all possible probabilities. However, it can be efficiently computed with the help of dynamic programming. During the model training, as we minimize the loss of the training dataset $D$, CTC loss is reformulated as the negative sum of log-likelihood. The equation is written as follows:

$$\sum_{(X,Y) \in D} -\log P(Y|X). \quad (2)$$

## 4 Experimental Setup

### 4.1 Description for Myanmar Speech Corpus

Due to the scarcity of public speech corpora for the Myanmar language, we build a corpus of read Myanmar speech. The dataset contains 9908 short audio clips each with a transcription. The content of the corpus is derived from the weather news of the "Department of Meteorology and Hydrology (Myanmar)" [18] and "DVB TVnews" [19].

The dataset contains 3 female and 1 male speakers. The audio clips are recorded in a quiet place with a minimum of background noise. The lengths of audio clips vary from 1 to 20 s and the total length of all audio clips is nearly 26 h. Audio clips are in single-channel 16-bit WAV format and are sampled at 22.05 kHz.

Most of the transcriptions are collected from the above-mentioned data sources and some are hand-transcribed. Since the transcriptions are mixed with both Zawgyi and Unicode font encodings, all the texts are firstly normalized into Unicode encoding. Myanmar Tools [20] is used to detect the Zawgyi strings and ICU Transliterator [21] is used to convert Zawgyi to Unicode. The numbers are also expanded into full words and the punctuations and the texts written in other languages are dropped. The audio clips and transcriptions are segmented and aligned manually.

For the experimental purpose, we randomly split 70% ($\sim$18 h) of the total data as the training set, 10% ($\sim$2 h) as the development set and 20% ($\sim$5 h) as the test set. We make sure to contain the audio clips of all four speakers in each dataset (Table 1).

Table 1. Statistics of the Myanmar speech corpus.

| | |
|---|---|
| Total clips | 9908 |
| Total duration | 25 h 18 min 37 s |
| Mean clip duration | 9.19 s |
| Min clip duration | 0.72 s |
| Max clip duration | 19.92 s |
| Mean character per clip | 121 |
| Mean syllable per clip | 38 |
| Distinct characters | 57 |
| Distinct syllables | 866 |

### 4.2 Training Setup

As input features, we use log spectrograms computed every 10 ms with 20 ms window. For CNN networks, the initial layers of the VGGNet with Batch Normalization layers are used as described in Sect. 3.1. The time resolution of the input features is reduced by 4 times after passing the convolutional layers. We use 3 to 6 layers of BLSTM each with 512 hidden units. A Batch Normalization with momentum 0.997 and epsilon 1e-5 is followed after each CNN and BLSTM layer. For the fully connected layer, we use the softmax activation function. All the models are trained with the CTC loss function. For the optimization, the Adam optimizer is used with the initial learning rate $\lambda$ of 1e-4.

The learning rate is reduced by the factor of 0.2 if the validation loss stops improving for certain epochs. The batch size of 8 is used in all experiments. The early stopping call back is also used to prevent overfitting. The different output label encoding methods are compared: character-level, syllable-level and sub-word level encodings. We use a set of regular expression rules [22] to segment the syllable-level tokens and use Byte Pair Encoding (BPE) to create the sub-word tokens. Since the experiments are performed with different label encoding methods, the results are evaluated using both CER and SER metrics. All the experiments are carried out on NVIDIA Tesla P100 GPU with 16GB GPU memory and 25 GB RAM (Google Colab). We do not use the external language model in this work. We will further investigate the effect of combining the decoding process with the external language model in the future.

## 5   Experimental Results

### 5.1   Effect of Using Different Label Encoding Methods

Only a small amount of research is done in character-level Myanmar ASR system. So it is quite interesting to explore Myanmar ASR models using the character-level output tokens. Since the Myanmar language is a syllable-timed language, the syllable-level tokens are also considered to be used. The dataset contains 57 unique characters and 866 unique syllables. We also conduct a number of experiments on the BPE sub-word tokenization in which a specific vocabulary size is needed to define. Because the dataset is relatively small, only small vocabulary sizes of 100, 300 and 500 BPE sub-word units are used in the experiments.

Table 2 shows the results of varying the different label encoding methods. It is observed that using large label encoding units is not very beneficial for a small dataset. The syllable-level model (866 tokens) and BPE model (500 tokens) show high error rates on both development and test sets. The best results are obtained by the character-level encoder with 4.69% CER and 11.86% SER on the development set and 4.72% CER and 12.38% SER on the test set.

### 5.2   Effect of Using Convolutional Layers

The use of many convolutional blocks may over-compress the number of features and make the length of the input audio features smaller than the length of output labels. This can cause problems in CTC calculation. This usually occurs when training the character-level ASR model of which the output of the model requires at least one time step per output character. For this reason, we limit the amount of downsampling with the maximum use of 2 max pooling layers.

As can be seen in Table 2, the use of convolutional layers tends to increase the performance of the model. All the models with convolutional layers outperform the models without convolutional layers. Moreover, the downsampling effect of the convolutional blocks significantly reduces the training time and speeds up the convergence of the model.

**Table 2.** Comparison of the effect of using convolutional layers and different label encoding units on both development and test sets. Each model contains the same number of convolutional layers, 5 layers of LSTM layers and a fully connected layer. The input features are not downsampled for the models without convolutional layers. The experiments are evaluated with both character error rate (CER) and syllable error rate (SER) metrics.

| Unit | With CNN | | | | Without CNN | | | |
|---|---|---|---|---|---|---|---|---|
| | Dev | | Test | | Dev | | Test | |
| | CER | SER | CER | SER | CER | SER | CER | SER |
| Char | **4.69** | **11.86** | **4.72** | **12.38** | **5.29** | **12.12** | **5.67** | **12.61** |
| Syllable | 21.95 | 20.67 | 22.68 | 22.34 | 22.09 | 23.46 | 23.41 | 24.83 |
| BPE 100 | 16.44 | 18.80 | 13.72 | 16.58 | 19.51 | 26.69 | 18.72 | 25.27 |
| BPE 300 | 9.61 | 19.12 | 10.38 | 20.35 | 9.65 | 20.88 | 9.98 | 21.83 |
| BPE 500 | 10.77 | 24.73 | 11.58 | 27.34 | 22.08 | 34.61 | 22.67 | 36.03 |

### 5.3 Effect of Using Different Numbers of BLSTM Layers

Since the Myanmar speech corpus is relatively small, the effect of varying model sizes are explored starting from a small model depth of 3. Only the depth of BLSTM layers is varied and the depth of CNN layers and other parameters are kept constant. Table 3 shows that the performance of the model keeps increasing until it reaches the depth of 6 LSTM layers on the test set. The model with 5 LSTM layers is the best fit for the small dataset which achieves the overall best results of 4.72% CER and 12.38% SER on test set.

**Table 3.** Comparison of character-level ASR models with different numbers of BLSTM layers. Convolutional layers are used in every experiment. A fully connected layer is followed after the BLSTM layers. The hidden unit size of BLSTM layers is set to 512 units.

| Number of BLSTM Layers | Dev | | Test | |
|---|---|---|---|---|
| | CER | SER | CER | SER |
| 3 | 4.65 | **11.85** | 5.03 | 13.24 |
| 4 | **4.58** | 12.54 | 4.93 | 13.75 |
| 5 | 4.69 | 11.86 | **4.72** | **12.38** |
| 6 | 5.26 | 14.45 | 5.45 | 15.46 |

## 6 Conclusion

In this paper, we explore the CTC based end-to-end architectures on the Myanmar language. We empirically compare the various label encoding methods, different depths of BLSTM layers and the use of convolutional layers on the low-resource Myanmar speech corpus. This work shows that a well-tuned end-to-end system can achieve state-of-the-art results in a closed domain ASR even for low-resource languages. As future work, we will further investigate the integration of the language model to our end-to-end ASR system and will explore the other end-to-end multi-tasking techniques.

**Acknowledgments.** The authors are grateful to the advisors from the University of Information Technology who gave us helpful comments and suggestions throughout this project. The authors also thank Ye Yint Htoon and May Sabal Myo for helping us with the dataset preparation and for technical assistance.